\documentclass[11pt]{jmlr}

\usepackage{booktabs}
\usepackage{listings}
\usepackage{xcolor}
\usepackage{bm}

\definecolor{codebg}{gray}{0.96}
\newcommand{\pkg}[1]{\texttt{#1}}
\newcommand{\Sone}{\Sigma_{1}}
\newcommand{\Sm}{\Sigma_{\mathrm{m}}}
\newcommand{\Sn}{\Sigma_{\mathrm{n}}}
\newcommand{\Sk}{\Sigma_{\mathrm{k}}}

\jmlrproceedings{preprint}{Preprint}
\jmlryear{2026}
\jmlrpages{}
\jmlrworkshop{}

\title[aggmetrics]{aggmetrics: Aggregating Classification Performance Measures
for Machine Learning}

\author{\Name{Ronaldo C.\ Prati} \Email{ronaldo.prati@ufabc.edu.br}\\
\addr Universidade Federal do ABC (UFABC), Santo Andr\'e, SP, Brazil}

\begin{document}

\maketitle

\begin{abstract}
\pkg{aggmetrics} is a Python package that extends any binary performance measure
expressed in the four confusion-matrix counts to multiclass, multilabel, ordinal,
cost-sensitive, soft-label and cross-validation settings, by substituting one of four
aggregation operators for what libraries usually expose as an \texttt{average}
switch. The package also stores, for every measure, the structural properties that
decide which identities apply to it, can infer those properties from a user-supplied
formula, and uses them to warn when a measure is used where it does not behave as
commonly assumed. Its test suite encodes the results of the companion theory as
executable tests. The package is BSD-3 licensed, depends only on NumPy, and is available at
\url{https://github.com/rcprati/aggmetrics}; documentation at
\url{https://aggmetrics.readthedocs.io}.
\end{abstract}

\section{Introduction}

Reporting a classifier's performance involves two choices that are made separately
in practice: which measure to use, and how to aggregate it over classes, examples or
cross-validation folds. In \pkg{scikit-learn} \citep{pedregosa2011scikit} the second
choice is an \texttt{average} argument attached to a fixed list of functions, so a
combination outside that list (a macro-averaged Matthews correlation coefficient, an
exemplar-averaged Jaccard index) has to be written by hand. More importantly, a
function returns a float whether or not the evaluation setting suits the measure.
Two examples reproducible with the package illustrate the problem.

\emph{Floors.} Under one-hot encoding of an $m$-class problem, each misclassified
example still contributes $m-2$ true negatives. On a 10-class problem, a classifier
that is wrong on every example scores $0.800$ on label accuracy and $-0.111$ on MCC
(micro-averaged), rather than $0$ and $-1$. Measures that depend on the true negatives
at $tp=0$ cannot fall below a floor that grows with $m$.

\emph{Micro-averaging.} Micro-averaged precision, recall and $F_1$ are all equal to
accuracy on single-label data, so reporting the three together conveys one number. And
for measures such as balanced accuracy or MCC, the pooled micro value can lie outside
the range of the per-class values, so no choice of weights reproduces it from a macro
average.

\pkg{aggmetrics} addresses both issues. It implements the construction of
\citet{prati2026unified}, in which micro-, macro-, exemplar- and fold-pooled
averaging are four instances of one operation, and it records the properties of each
measure that determine such behaviour, so they can be queried and checked.

\section{Design}

\paragraph{Construction.}
Actual and predicted labels are held as indicator matrices
$\bm{Y},\hat{\bm{Y}}\in\{0,1\}^{n\times m}$ (real-valued for soft labels; for
ordinal data with $m$ classes, the $n\times(m-1)$ matrix of cumulative thresholds
$\mathbb{I}(y_i>j)$). Four cell matrices $\bm{TP}=\bm{Y}\wedge\hat{\bm{Y}}$, and
analogously $\bm{FP}$, $\bm{FN}$, $\bm{TN}$, are reduced by an operator:
$\Sone$ over all entries (micro), $\Sm$ over columns (macro), $\Sn$ over rows
(exemplar), and $\Sk$ across cross-validation splits. Substituting the reduced counts
into a binary measure $M(tp,fp,fn,tn)$ gives its extension. The 21 registered measures
and four operators therefore give 84 combinations through one code path, with
\pkg{scikit-learn}'s \texttt{'micro'}, \texttt{'macro'}, \texttt{'weighted'} and
\texttt{'samples'} as a subset. Constructors accept class-index vectors, binary
matrices, probability matrices and ordinal classes; the cells are cached, so a report
over many measures costs one pass.

\paragraph{Structural properties.}
Each measure stores whether it is \emph{aggregation-decomposable}
($M=\langle\bm a,\bm v\rangle/\langle\bm b,\bm v\rangle$ with $\bm v=(tp,fp,fn,tn)$,
which is the condition under which micro equals denominator-weighted macro), whether it
is skew-invariant, whether its value at $tp=0$ depends on $tn$, and its direction.
Bayes decision thresholds and attainable floors follow in closed form. Properties are
data, not documentation:

\begin{lstlisting}
get_measure('mcc').decomposable          # False
get_measure('recall').bayes_threshold()  # 0.0: maximised by predicting all positive
get_measure('mcc').attainable_floor(10)  # -0.111
\end{lstlisting}

\paragraph{Inference from a formula.}
\texttt{register\_measure} accepts any callable of $(tp,fp,fn,tn)$ and derives the
properties above: decomposability by solving a homogeneous linear system for
$(\bm a,\bm b)$ over sampled count vectors and validating it on held-out vectors,
and the remaining properties by evaluation. The new measure is then usable under every operator.

\begin{lstlisting}
def prevalence_threshold(tp, fp, fn, tn):
    r, s = tp / (tp + fn), tn / (tn + fp)
    return (np.sqrt(r * (1 - s)) + s - 1) / (r + s - 1)

m = register_measure('prevalence_threshold', prevalence_threshold)
m.decomposable      # False (inferred)
m.skew_invariant    # True
im.compute('prevalence_threshold', 'exemplar')
\end{lstlisting}

\noindent
Because the 21 built-in measures have analytically derived properties, inference can
be checked against them: \texttt{verify\_inference()} agrees on all 21, and the recovered
coefficients match the analytic ones to within $6.1\times10^{-15}$, about $28$ times
the double-precision machine epsilon ($\varepsilon=2.2\times10^{-16}$). The check found two errors in
our own hand annotations (the error rate and the false positive rate both retain a
$tn$ dependence at $tp=0$, so both flatter a classifier that is wrong everywhere).

\paragraph{Diagnostics.}
\texttt{diagnose(im)} returns a list of \texttt{Finding} objects, each with a code, a
severity, a message and the triggering values. The checks cover: pooled micro value
outside the per-class range (\texttt{outside\_hull}); $tn$-dependent measures on one-hot
data (\texttt{inflated\_floor}); micro-collapse on single-label data
(\texttt{micro\_collapse}); undefined per-class entries and the size of the
convention effect (\texttt{undefined\_entries}); measures whose optimum is a trivial or
non-Bayes rule (\texttt{degenerate\_optimum}, \texttt{not\_erm\_consistent});
one-vs-rest prior drift (\texttt{ovr\_skew\_drift}); and evaluation suites that exceed the $m(m-1)$
degrees of freedom of an $m$-class confusion matrix (\texttt{redundant\_suite}).
With \texttt{emit\_warnings=True} findings are raised as Python warnings, so
an evaluation script can fail in continuous integration on a flawed protocol.

\paragraph{Further functionality.}
Four conventions for $0/0$ entries (\texttt{exclude}, \texttt{zero}, \texttt{one},
\texttt{nan}) are an explicit parameter; \texttt{zero} reproduces
\pkg{scikit-learn}'s \texttt{zero\_division=0}. Soft ground truth uses t-norms in place
of conjunction; \texttt{marginal\_defect} shows that only the product t-norm preserves
the marginals. Cost matrices recover MAE and MSE for $C_{jk}=|j-k|$ and $(j-k)^2$.
Observation masks restrict every operator to observed entries, covering partial labels
and abstention. \texttt{pareto\_dominance(im\_a, im\_b)} decides whether one model beats
another on all monotone measures, and \texttt{make\_scorer} yields scorers usable
in \texttt{cross\_val\_score} and \texttt{GridSearchCV}.

\section{Example}

\begin{lstlisting}
import numpy as np
from aggmetrics import IndicatorMatrix, diagnose

y    = np.arange(500) % 10
yhat = (y + 1) % 10                       # wrong on every example
im = IndicatorMatrix(y, yhat, task='multiclass')

im.accuracy()                             # 0.000
im.compute('label_accuracy', 'micro')     # 0.800
im.compute('mcc', 'micro')                # -0.111
for f in diagnose(im):
    print(f)
# [warning] inflated_floor: 'balanced_accuracy' cannot fall below 0.4444 on this
#           10-class one-hot problem even if every prediction is wrong ...
# [note]    micro_collapse: micro-precision, micro-recall and micro-F1 are all
#           equal to accuracy ...
\end{lstlisting}

\noindent
In a model-selection loop, the scorer interface applies the same measures:

\begin{lstlisting}
from aggmetrics import make_scorer
from sklearn.model_selection import cross_val_score
cross_val_score(clf, X, y, cv=5, scoring=make_scorer('mcc', 'macro', task='multiclass'))
\end{lstlisting}

\section{Comparison with existing software}

\paragraph{\pkg{scikit-learn} and \pkg{torchmetrics}.}
\pkg{scikit-learn} \citep{pedregosa2011scikit} and \pkg{torchmetrics}
\citep{detlefsen2022torchmetrics} provide accurate, widely used implementations of a
fixed catalogue of measures with an averaging argument. \pkg{aggmetrics} does not
replace them for production metric computation (\pkg{torchmetrics} targets GPU,
streaming and distributed accumulation, which \pkg{aggmetrics} does not); it offers the full
measure--operator grid, explicit undefined-value conventions, and the diagnostic
layer. Its scorers interoperate with \pkg{scikit-learn}; for $F_1$ on a random
multilabel problem the four \pkg{scikit-learn} averages are reproduced to machine
precision (with \texttt{convention='zero'} for the exemplar average).
\pkg{imbalanced-learn} \citep{lemaitre2017imbalanced} adds measures for imbalanced data, such as the geometric mean
and index balanced accuracy, again as a fixed list.

\paragraph{\pkg{PyCM} and \pkg{yardstick}.}
\pkg{PyCM} \citep{haghighi2018pycm} derives a large set of per-class and overall
statistics from a confusion matrix and is the closest Python package in breadth of measures; it takes a
confusion matrix as input, so multilabel and soft-label settings are not
representable, and it does not offer user-defined measures with derived properties.
The R package \pkg{yardstick} \citep{kuhn2024yardstick} supports macro, weighted
macro and micro estimators for multiclass metrics. To our knowledge none of these
packages attaches structural properties to measures or warns about floor effects, but we have
not benchmarked their behaviour on the cases above beyond reading their documentation.

\section{Project status}

\paragraph{Licence and dependencies.}
BSD 3-Clause. The core requires only NumPy ($\geq1.20$); \pkg{scikit-learn} and \pkg{pandas}
are optional (scorers and tabular reports). Pure Python, no compilation, no proprietary
dependencies; Python $\geq3.9$. About 1800 lines of source.

\paragraph{Testing.}
The suite has 209 tests and runs in about three seconds. Each theoretical
result of the companion paper is an executable test
(e.g.\ micro~$=$~denominator-weighted macro for the decomposable measures,
uniqueness of the product t-norm, micro-collapse, Bayes thresholds, attainable floors,
recovery of MAE/MSE from cost matrices), so a change that breaks a theorem
breaks a test. Line and branch coverage are both 100\%, and continuous
integration (GitHub Actions) fails if they drop. It runs
the tests, the doctests and the worked example on Linux, macOS and Windows with Python
3.9--3.12.

\paragraph{Documentation and maintenance.}
The repository contains a README with a feature tour, a runnable
\texttt{examples/quickstart.py}, docstrings on all public functions, and a machine-readable
guide for AI coding assistants (\texttt{.agents/skills/}). Version~0.1.0 is the version under review.
It is archived on Zenodo (\texttt{10.5281/zenodo.23535051}), and the documentation, with a
gallery of example notebooks, is at \url{https://aggmetrics.readthedocs.io}. A \texttt{CITATION.cff} is provided.

\section{Conclusion}
\pkg{aggmetrics} makes the aggregation operator an explicit parameter for every measure,
keeps the properties that determine which identities hold alongside the measures, and
uses them to flag misleading evaluations. Planned work includes sparse backends
for large label spaces, a streaming accumulator, and an R implementation.

\acks{This research was partially supported by CNPq and FAPESP, Brazil.}

\clearpage
\bibliography{aggmetrics_mloss}

\end{document}